\def\year{2020}\relax
\documentclass[letterpaper]{article} 
\usepackage{aaai20}  
\usepackage{times}  
\usepackage{helvet} 
\usepackage{courier}  
\usepackage[hyphens]{url}  
\usepackage{graphicx} 
\usepackage{graphicx}  
\usepackage{subcaption}
\usepackage{amsmath} 
\usepackage{amsfonts}

\title{Learning Agent Communication under Limited Bandwidth by Message Pruning}
\author{Hangyu Mao,\textsuperscript{\rm 1}
	Zhengchao Zhang,\textsuperscript{\rm 1}
	Zhen Xiao,\textsuperscript{\rm 1}
	Zhibo Gong,\textsuperscript{\rm 2}
	Yan Ni\textsuperscript{\rm 1}\\
	\textsuperscript{\rm 1}Peking University,
	\textsuperscript{\rm 2}Huawei Technologies Co., Ltd.\\
	\{hy.mao, zhengchaozhang, xiaozhen, niyan.ny\}@pku.edu.cn, gongzhibo@huawei.com
}

\begin{document}

\maketitle

\begin{abstract}
Communication is a crucial factor for the big multi-agent world to stay organized and productive. Recently, Deep Reinforcement Learning (DRL) has been applied to learn the communication strategy and the control policy for multiple agents. However, the practical \emph{\textbf{limited bandwidth}} in multi-agent communication has been largely ignored by the existing DRL methods. Specifically, many methods keep sending messages incessantly, which consumes too much bandwidth. As a result, they are inapplicable to multi-agent systems with limited bandwidth. To handle this problem, we propose a gating mechanism to adaptively prune less beneficial messages. We evaluate the gating mechanism on several tasks. Experiments demonstrate that it can prune a lot of messages with little impact on performance. In fact, the performance may be greatly improved by pruning redundant messages. Moreover, the proposed gating mechanism is applicable to several previous methods, equipping them the ability  to address bandwidth restricted settings.
\end{abstract}

\section{Introduction}
Communication is an essential human intelligence. It is a critical factor to keep the world in order. Recently, inspired by the communication among humans, Deep Reinforcement Learning (DRL) has been adopted to learn the communication among multiple artificial agents.

However, it remains an open question to apply the existing methods to real-world multi-agent systems because real-world applications usually impose many constraints on communication, e.g., the bandwidth limitation for transmitting messages and the protection of private messages. In order to develop practical communication strategies, these constraints need to be resolved.

In this paper, we focus on addressing the \textbf{\emph{limited bandwidth}} problem in multi-agent communication. As we know, the bandwidth (and more generally, the resource) for transmitting the communication messages is limited \cite{roth2005reasoning,roth2006communicate,becker2009analyzing,wu2011online,zhang2013coordinating}. Thus, the agents should generate as few messages as possible on the premise of maintaining the performance.

We are interested in the limited bandwidth problem due to two reasons. On the one hand, it is ubiquitous in the real world. For example, in the wired packet routing systems, the links have a limited transmission capacity; in the wireless Internet of Things, the sensors have a limited battery capacity. Once we figure out a principled method to address this problem, many fields may benefit from this work. On the other hand, this problem has been largely ignored by the existing DRL methods. There is a great need to devote attention to this problem.

We take two steps to address this problem. Firstly, we aggregate the merits of the existing methods to form a basic model named \emph{Actor-Critic Message Learner} (ACML). However, the basic ACML is still not practical because it does not change the communication pattern of the existing methods. That is to say, ACML and many previous methods send messages incessantly, regardless of whether the message is beneficial for the agent team. Secondly, we extend ACML with a gating mechanism to design a more flexible and practical Gated-ACML model. The gating mechanism is trained based on a novel auxiliary task, which tries to open the gate to encourage communication when the message is beneficial enough for the agent team, and close the gate to discourage communication otherwise. As a result, after the gating mechanism has been well-trained, it can prune unbeneficial messages adaptively to control the message quantity around a desired threshold. Consequently, Gated-ACML is applicable to multi-agent systems with limited bandwidth.

Our contributions are summarized as follows.

(1) We are the first to formally study the limited bandwidth problem using DRL. We give the first detailed analyses about how gating mechanism can serve this purpose.

(2) We apply the gating mechanism to several previous methods, equipping them the ability to address bandwidth restricted settings.

(3) We evaluate the gating mechanism on several simulators driven by the real-world tasks. Experiments show that it can prune a lot of messages with little impact on performance, or with great improvement on performance in specific scenarios. These are very fundamental findings in the community. It could be the best contribution of our work beyond the gating mechanism.

\section{Related Work} \label{sec:RelatedWork}
Communication is vital for multi-agent systems. Recently, the communication channel implemented by Deep Neural Network (DNN) has been proven useful for learning beneficial messages \cite{sukhbaatar2016learning,foerster2016learning,peng2017multiagent,peng2018learning,mao2017accnet,kong2017revisiting,mao2019modelling,Lowe2017Multi,Chu2017Parameter,kilinc2018multi,kim2019learning,jiang2018learning,singh2018learning,kim2019message}.

CommNet \cite{sukhbaatar2016learning} embeds a centralized communication channel into the actor network, and it processes other agents' messages by averaging them. AMP \cite{peng2018learning} adopts a similar design, but it applies an attention mechanism to focus on specific messages from other agents. Relevant studies include but are not limited to \cite{foerster2016learning,mao2017accnet,kong2017revisiting}. However, the policy networks used in these methods take as input the messages from all agents to generate a single control action. Thus, the agents have to keep sending messages incessantly in every control cycle, without alternatives to reduce the message quantity. Due to emitting too many messages, they are inflexible to be applied to multi-agent systems with limited bandwidth.

We argue that the principled way to address the limited bandwidth problem is to apply some special mechanisms to adaptively decide whether to send (equivalently, whether to \emph{prune}) the current message. From this perspective, the most relevant studies to our Gated-ACML are IC3Net \cite{singh2018learning} and ATOC \cite{jiang2018learning}. Technically, the three methods adopt the so-called gating mechanism to generate a binary action to specify whether the agent should communicate with others. The fundamental difference is the research purpose: IC3Net aims to learn when to communicate at scale in both cooperative and competitive tasks; ATOC targets at learning to communicate with a dynamic agent topology; as far as we know, Gated-ACML is the first formal DRL method to address the limited bandwidth problem.

As a result of different research purposes, the implementations are very different. IC3Net allows multiple communication cycles in one step, and it generates the gating value for the next communication cycle. ATOC relies on the gating value to form dynamic local communication group, and it may classify one agent into many agent groups. Because these implementations do not explicitly consider bandwidth limitation, they may generate a lot of messages (as indicated by the experiments). In contrast, Gated-ACML adopts a global Q-value difference as well as a specially designed threshold to identify the beneficial messages, and it applies the gating value to prune the current messages of a global communication group. Therefore, it has more sophisticated ability to control how many messages will be pruned and to satisfy the needs of different applications.

Meanwhile, IC3Net and ATOC are only suitable for homogeneous agents due to their DNN structures, and they are designed for discrete action. By contrast, Gated-ACML is suitable for both homogeneous and heterogeneous agents, and it is designed for continuous action.

\section{Background}
\textbf{DEC-POMDP.} We consider a \emph{fully cooperative} multi-agent setting that can be formulated as DEC-POMDP \cite{bernstein2002complexity}. It is formally defined as a tuple $\langle N,S,\vec{A},T,R,\vec{O},Z,\gamma \rangle$, where $N$ is the number of agents; $S$ is the set of state $s$; $\vec{A}=[A_1, ..., A_N]$ represents the set of \emph{joint action} $\vec{a}$, and $A_i$ is the set of \emph{local action} $a_i$ that agent $i$ can take; $T(s'|s,\vec{a}): S \times \vec{A} \times S \rightarrow [0,1]$ represents the state transition function; $R: S \times \vec{A} \times S \rightarrow \mathbb{R}$ is the reward function shared by all agents; $\vec{O} = [O_1, ..., O_N]$ is the set of \emph{joint observation} $\vec{o}$ controlled by the observation function $Z: S \times \vec{A} \rightarrow \vec{O}$; $\gamma \in [0,1]$ is the \emph{discount factor}.

In a given state $s$, agent $i$ can only observe an observation $o_i$, and each agent takes an action $a_i$ based on its observation $o_i$ (and possibly messages from other agents), resulting in a new state $s'$ and a shared reward $r$. The agents try to learn a policy $\pi_{i}(a_i|o_i): O_i \times A_i \rightarrow [0,1]$ that can maximize $\mathbb{E}[G]$ where $G$ is the \emph{discount return} defined as $G = \sum_{t=0}^{H} \gamma^{t} r^{t}$, and $H$ is the time horizon. In practice, we map the observation history rather than the current observation to an action, namely, $o_i$ represents the observation history of agent $i$ in the rest of the paper.

\textbf{Reinforcement Learning (RL).} RL \cite{sutton1998introduction} is generally used to solve special DEC-POMDP problems where $N=1$. In practice, we define the Q-value (or action-value) function as $Q^{\pi}(s,a) = \mathbb{E}_{\pi}[G|S=s,A=a]$, then the optimal policy $\pi^{*}$ can be derived by $\pi^{*} = \arg\max_{\pi} Q^{\pi}(s,a)$. The actor-critic algorithm is one of the most effective RL methods because the actor and the critic can reinforce each other, so we design our Gated-ACML based on actor-critic method. Deterministic Policy Gradient (DPG) \cite{silver2014deterministic} is a special actor-critic algorithm where the actor adopts a deterministic policy $\mu_{\theta}: S \rightarrow A$ and the action space $A$ is continuous. Deep DPG (DDPG) \cite{lillicrap2015continuous} applies DNN $\mu_{\theta}(s)$ and $Q(s,a;w)$ to represent the actor and the critic, respectively. DDPG is an off-policy method. It adopts \emph{target network} and \emph{experience replay} to stabilize training and to improve data efficiency. Specifically, the critic's parameters $w$ and the actor's parameters $\theta$ are updated based on:
\begin{eqnarray}
L(w) =& \hspace{-11.5em} \mathbb{E}_{(s,a,r,s') \sim D}[\delta^{2}] \label{equ:DPG1} \\
\delta =& \hspace{-0.5em} r + \gamma Q(s',a';w^{-})|_{a'=\mu_{\theta^{-}}(s')} - Q(s,a;w) \label{equ:DPG2} \\
\nabla_{\theta}J(\theta) =& \hspace{-1.9em} \mathbb{E}_{s \sim D}[\nabla_{\theta}\mu_\theta(s) * \nabla_{a}Q(s,a;w)|_{a=\mu_{\theta}(s)}] \label{equ:DPG3}
\end{eqnarray}
where $L(w)$ is the loss function of the critic; $J(\theta)$ is the objective function of the actor; $D$ is the replay buffer containing recent experience tuples $(s,a,r,s')$; $Q(s,a;w^{-})$ and $\mu_{\theta^{-}}(s)$ are the target networks whose parameters $w^{-}$ and $\theta^{-}$ are periodically updated by copying $w$ and $\theta$. A merit of actor-critic algorithms is that the critic is only used during training, and it will be naturally removed during execution, so we only need to prune messages among the actors in our Gated-ACML.

\section{Methods}
We list the key notations as follows. $a_i$ is the local action of agent $i$. $\vec{a}_{-i}$ is the joint action of other agents except for agent $i$. $\vec{a}$ is the joint action of all agents, i.e., $\vec{a}=\langle a_{i}, \vec{a}_{-i} \rangle$. The observation history $\vec{o}$, $o_i$, $\vec{o}_{-i}$, and the policy $\mu_{\theta_i}$ are denoted similarly. $s'$ is the next state after $s$, and $\vec{o}'$, $o_i'$, $\vec{o}_{-i}'$, $\vec{a}'$, $a_i'$, $\vec{a}_{-i}'$ are denoted in a similar sense.

\subsection{ACML}
\begin{figure}[!htb]
	\centering
	\includegraphics[width=.95\columnwidth]{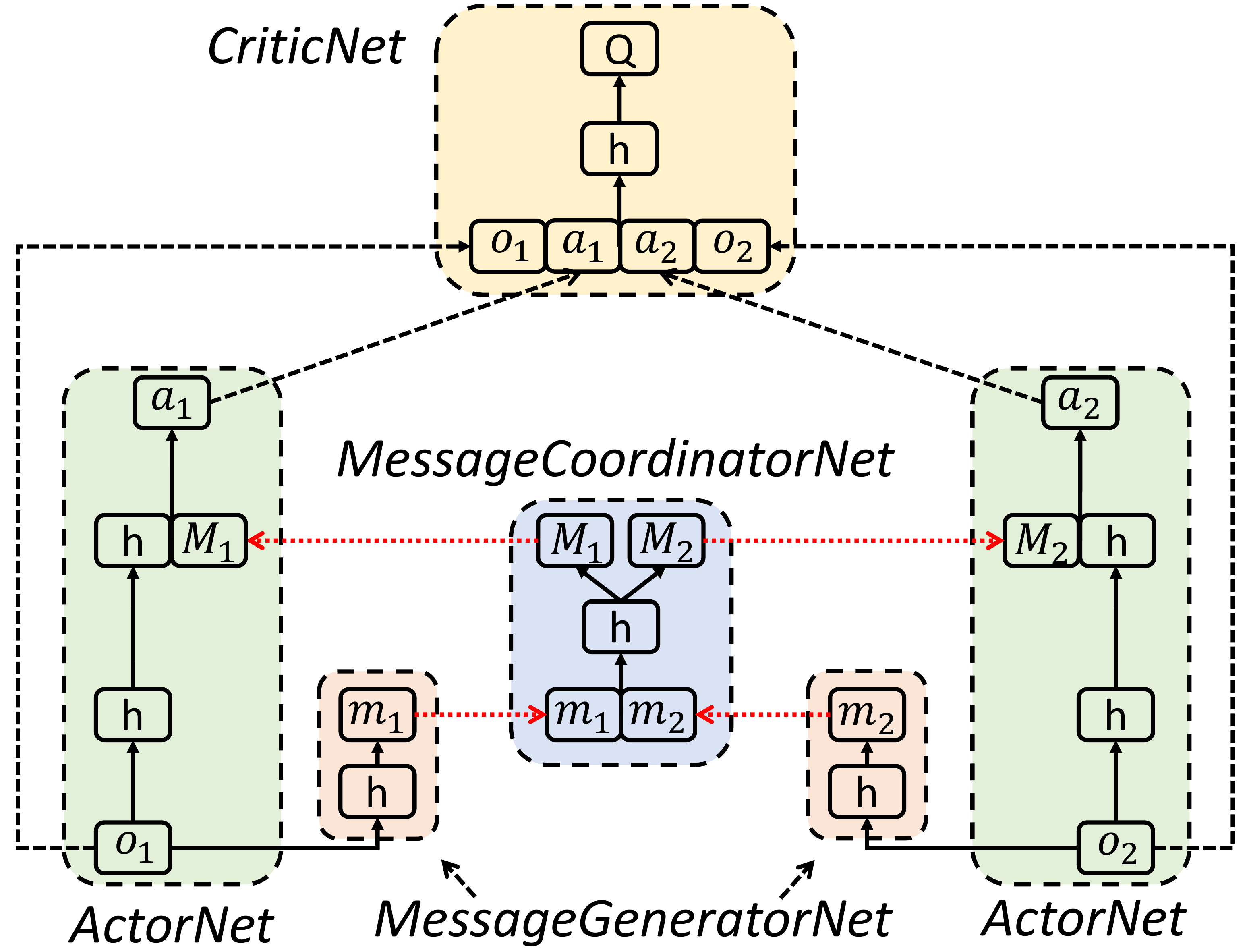}
	\caption{The proposed ACML. For clarity, we illustrate this model with a two-agent example. All components are implemented by DNN. $h$ is the hidden layer of the DNN; $m_i$ is the local message; $M_i$ is the global message. The red arrows indicate the message exchange process.}
	\label{fig:MODEL_ACML}
\end{figure}

\textbf{Design.} ACML is motivated by combining the merits of the existing methods. As Figure \ref{fig:MODEL_ACML} shows, ACML adopts the following design: each agent is composed of an ActorNet and a MessageGeneratorNet, while all agents share the same CriticNet and MessageCoordinatorNet. All components are implemented by DNN. The shared MessageCoordinatorNet is similar to the communication channel of many previous methods such as CommNet, AMP and ATOC, while the shared CriticNet is the same as the well-known MADDPG \cite{Lowe2017Multi}. By sharing the encoding of observation and action within the whole agent team, individual agents could build up relatively greater global perception, infer the intent of other agents, and cooperate on decision making \cite{jiang2018learning}.

In addition, aggregating these components properly could relieve the problems encountered by previous methods. For example, MADDPG and \cite{mao2019modelling,Chu2017Parameter} do not adopt the MessageCoordinatorNet, making them suffer from the partially observable problem; while AMP and ATOC do not adopt the CriticNet, making them suffer from the non-stationary problem during training \cite{hernandez2017survey}. In contrast, ACML is fully observable and training stationary due to the careful design.

ACML works as follows during execution. (1) $m_i = \text{MessageGeneratorNet}(o_i)$, i.e., agent $i$ generates the local message $m_i$ based on its observation $o_i$. (2) All agents send the message $m_i$ to the MessageCoordinatorNet. (3) $M_1,.., M_N = \text{MessageCoordinatorNet}(m_1,.., m_N)$, i.e., the MessageCoordinatorNet extracts the global message $M_i$ for each agent $i$ based on all local messages. (4) The MessageCoordinatorNet sends $M_i$ back to agent $i$. (5) $a_i = \text{ActorNet}(o_i, M_i)$, i.e., agent $i$  generates action $a_i$ based on its local observation $o_i$ and the global message $M_i$.

\textbf{Training.} The agents generate $a_i$ based on $o_i$ and $M_i$ to interact with the environment, and the environment will feed a shared reward $r$ to the agents. Then, the experience tuples $\langle o_i, \vec{o}_{-i}, a_i, \vec{a}_{-i}, r, o'_i,\vec{o}'_{-i} \rangle$ are used to train ACML. Specifically, as the agents exchange messages with each other, the actor and the shared critic can be represented as $\mu_{\theta_i}(o_i,M_i)$ and $Q(\vec{o},\vec{a};w)$, respectively. We can extend Equation (\ref{equ:DPG1} -- \ref{equ:DPG3}) to multi-agent formulations as follows:
\begin{eqnarray}
L(w) =& \hspace{-6.8em} \mathbb{E}_{(o_i,\vec{o}_{-i},a_i,\vec{a}_{-i},r,o'_i,\vec{o}'_{-i}) \sim D}[\delta^{2}] \label{equ:mDPG1} \\
\delta =& \hspace{-0.8em} r+\gamma Q(\vec{o}', \vec{a}';w^{-})|_{a'_i=\mu_{\theta_i^{-}}(o'_i)} - Q(\vec{o}, \vec{a};w) \label{equ:mDPG2} \\
\nabla_{\theta_i}J(\theta_i) =& \hspace{-1.0em} \mathbb{E}_{(o_i,\vec{o}_{-i}) \sim D}[ \nabla_{\theta_i}\mu_{\theta_i}(o_i, M_i) \nabla_{a_i}Q(\vec{o}, \vec{a};w)] \label{equ:mDPG3}
\end{eqnarray}
where $L(w)$, $J(\theta_i)$, $w$, $\theta_i$, $w^{-}$ and $\theta_i^{-}$ have similar meanings as the single-agent setting. As for the gradient to update the parameters $\theta^{cc}$ of communication channel (i.e., the MessageGeneratorNet and MessageCoordinatorNet), it can be further derived by the chain rule as:
\begin{eqnarray}
\mathbb{E}_{(o_i,\vec{o}_{-i}) \sim D ,\hspace{0.1em} i \sim [1, N]} [ \nabla_{\theta^{cc}}M_i(m_1, ..., m_N;\theta^{cc}) * \nonumber \\
\nabla_{M_i}\mu_{\theta_i}(o_i, M_i) * \nonumber \\
\nabla_{a_i}Q(\vec{o}, \vec{a};w)|_{a_i=\mu_{\theta_i}(o_i)} ] \label{equ:CommunicationChannel}
\end{eqnarray}
Since ACML is end-to-end differentiable, the communication message and the control policy can be optimized jointly using back-propagation based on the above equations.

\subsection{Gated-ACML} \label{sec:Gated-ACML}
\textbf{Motivation.} As the execution process shows, ACML takes as input the encoding of messages from all agents to generate a control action. The agents have to send messages incessantly, regardless of whether the messages are beneficial to the performance of the agent team. This is a common problem of many previous methods such as CommNet \cite{sukhbaatar2016learning}, AMP \cite{peng2018learning} and \cite{foerster2016learning,mao2017accnet,kong2017revisiting}.
As a result, these methods usually consume a lot of bandwidths and resources, and they are unpractical for multi-agent systems with limited bandwidth.

\begin{figure}[!htb]
	\centering
	\includegraphics[width=.9\columnwidth]{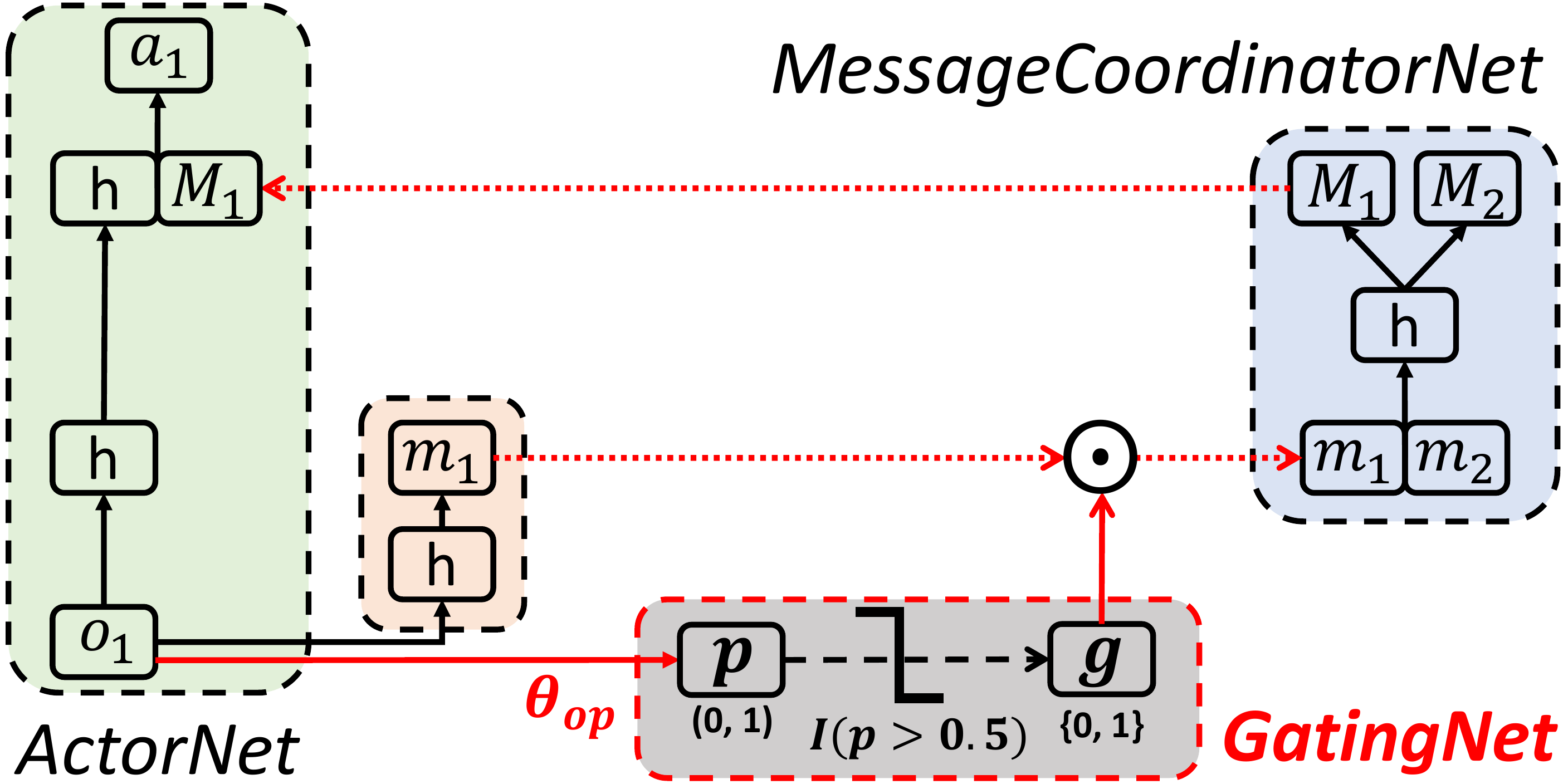}
	\caption{The actor part of the proposed Gated-ACML. For clarity, we only show one agent's structure, and we do not show the critic part because it is the same as that of ACML.}
	\label{fig:MODEL_ACML_Gated}
\end{figure}

We argue that the principled way to address this problem is to apply some special mechanisms to adaptively decide whether to \textbf{\emph{prune}} the messages. Gated-ACML is motivated by this idea, and it adopts a gating mechanism to adaptively prune less beneficial messages among ActorNets, such that the agents can maintain the performance with as few messages as possible.

\textbf{Design.} As shown in Figure \ref{fig:MODEL_ACML_Gated}, besides the original components, each agent is equipped with an additional GatingNet. Specifically, Gated-ACML works as follows. (1) The agent generates a local message $m_i$ as well as a probability score $p \in (0, 1)$ based on the observation $o_i$. (2) A gate value $g \in \{0, 1\}$ is generated by the indicator function $g \leftarrow \mathbb{I}(p>0.5)$. That is to say, if $p>0.5$, we set $g=1$, otherwise, we set $g=0$. (3) The agent sends $m_i \odot g$ to the MessageCoordinatorNet, and the following process is the same as that of ACML.

As can be figured out, if $g=1$, $m_i \odot g$ equals to the original message $m_i$, thus Gated-ACML is equivalent to ACML. In contrast, if $g=0$, $m_i \odot g$ will be a zero vector, which means that the message can be pruned. In practice, if the agent has not received a message when the decision making time arrived, it will pad zeros automatically. This design is supported by the message-dropout technique \cite{kim2019message}, which has shown that replacing some messages by zero vectors with a certain probability can make the training robust against communication errors and message redundance that are common for large-scale communication \cite{kilinc2018multi,kim2019learning,jiang2018learning,singh2018learning,kim2019message}. We find similar phenomena in our experiments.

\textbf{Training with Auxiliary Task.} In order to make the above design work, a suitable probability $p$ must be trained for each observation $o_i$, otherwise Gated-ACML may degenerate to ACML in the extreme case where $\mathbb{I}(p>0.5)$ always equals to 1. However, as the indicator function $g \leftarrow \mathbb{I}(p>0.5)$ is non-differentiable, it makes the end-to-end back-propagation method inapplicable. We have tried approximate gradient \cite{Hubara2016Binarized}, sparse regularization \cite{Makhzani2015Winner} and several other methods without success. To bypass the training of the non-differentiable indicator function, we decide to train its input $p$ directly. To this end, we adopt the auxiliary task technique \cite{Jaderberg2016Reinforcement} to provide training signal for $p$ explicitly.

Recall that we want to prune the messages on the premise of maintaining the performance. Because the performance of RL could be measured by the Q-value, i.e., the expected long term cumulative rewards, we design the following auxiliary task. Let $p$ indicate the probability that $\Delta Q(o_i) = Q(\langle o_{i}, a_{i}^{C} \rangle, \langle \vec{o}_{-i}, \vec{a}_{-i}^{C} \rangle) - Q(\langle o_{i}, a_{i}^{I} \rangle, \langle \vec{o}_{-i}, \vec{a}_{-i}^{C} \rangle)$ is larger than $T$, where $a_i^{C}$ is the action generated based on communication, $a_i^{I}$ is the action generated independently (i.e., without communication), and $T$ is a threshold controlling how many messages should be pruned. In this setting, the label of this auxiliary task can be formulated as:
\begin{eqnarray}
Y(o_i) = \mathbb{I}(\Delta Q(o_i) > T) = \mathbb{I}(Q(\langle o_{i}, a_{i}^{C} \rangle, \langle \vec{o}_{-i}, \vec{a}_{-i}^{C} \rangle) \nonumber \\
- Q(\langle o_{i}, a_{i}^{I} \rangle, \langle \vec{o}_{-i}, \vec{a}_{-i}^{C} \rangle) > T) \label{equ:AuxiliaryLabel}
\end{eqnarray}
where $\mathbb{I}$ is the indicator function. Then we can train $p$ by minimizing the following loss function:
\begin{eqnarray}
L_{\theta_{op}}(o_i) = - \mathbb{E}_{o_i}[ \hspace{0.3em} Y(o_i) \log p(o_i|\theta_{op}) \nonumber \\
& \hspace{-8.8em} + (1-Y(o_i)) \log (1-p(o_i|\theta_{op})) \hspace{0.3em} ] \label{equ:LossFunctionGating}
\end{eqnarray}
where $\theta_{op}$ is the parameters between the observation $o_i$ and the probability $p$ as shown in Figure \ref{fig:MODEL_ACML_Gated}.

Equation \ref{equ:AuxiliaryLabel} implies that only agent $i$ changes the behaviors between $a_{i}^{C}$ and $a_{i}^{I}$ to calculate $\Delta Q(o_i)$, while other agents are fixed to send messages all the time (i.e., keep $\vec{a}_{-i}^{C}$ unchanged) during training. To make sure that all agents can be trained properly, we train each agent based on the above equations in a round-robin manner, namely, $i$ cycles from $1$ to $N$ and to $1$ again. From the perspective of optimization, our method can be analogous to coordinate descent \footnote{\url{https://en.wikipedia.org/wiki/Coordinate_descent}} if we take the agents as coordinates, since we optimize a specific coordinate hyperplane while fixing other coordinates (i.e., optimize agent $i$ while fixing other agents \footnote{The assumption that $\vec{a}_{-i}^{C}$ keeps unchanged may be inconsistent with the reality because other agents $j$ may take $\vec{a}_{j}^{I}$ instead of $\vec{a}_{j}^{C}$. Nevertheless, it is a lower bound case of our optimization objective. In practice, coordinate descent can optimize this lower bound with high efficiency in huge-scale multi-agent setting \cite{Nesterov2012Efficiency}.}).

The insight of Equation \ref{equ:LossFunctionGating} is that if the label of the auxiliary task is $Y(o_i)=1$, namely, $a_i^{C}$ can really obtain at least $T$ Q-values more than $a_i^{I}$, the network should try to generate a probability $p(o_i;\theta_{op})$ that is larger than $0.5$ to encourage communication (recall that $g \leftarrow \mathbb{I}(p>0.5)$). In other words, Gated-ACML discourages communication by pruning the messages that contribute smaller Q-values than the threshold $T$. Therefore, after the gating mechanism has been well-trained, it can prune less beneficial messages adaptively to control the message quantity around a desired threshold specified by $T$. This is our key contribution beyond previous communication methods.

\textbf{Key Implementation.}
The training method relies on correct label of the auxiliary task, so we should provide suitable $Q(\langle o_{i}, a_{i}^{C} \rangle, \langle \vec{o}_{-i}, \vec{a}_{-i}^{C} \rangle)$, $Q(\langle o_{i}, a_{i}^{I} \rangle, \langle \vec{o}_{-i}, \vec{a}_{-i}^{C} \rangle)$ and $T$ as indicated by Equation \ref{equ:AuxiliaryLabel}.

For the Q-values, we firstly set $g=1$ (i.e., without message pruning) to train other components except for the GatingNet based on Equation (\ref{equ:mDPG1} -- \ref{equ:CommunicationChannel}). After the model is trained well, we can get approximately correct ActorNet and CriticNet. Then, for a specific $o_i$, the ActorNet can generate $a_{i}^{C}$ and $a_{i}^{I}$ when we set $g=1$ and $g=0$, respectively. Afterwards, taking $o_i$, $o_{-i}$ as well as the generated $a_{i}^{C}$, $a_{i}^{I}$ and $\vec{a}_{-i}^{C}$ as input, the CriticNet can estimate approximately correct Q-values $Q(\langle o_{i}, a_{i}^{C} \rangle, \langle \vec{o}_{-i}, \vec{a}_{-i}^{C} \rangle)$ and $Q(\langle o_{i}, a_{i}^{I} \rangle, \langle \vec{o}_{-i}, \vec{a}_{-i}^{C} \rangle)$.

For the threshold $T$, we propose two methods to set a fixed $T$ and a dynamic $T$, respectively. For a fixed $T$, we firstly sort the $\Delta Q(o_i)$ of the latest $K$ observations $o_i$ encountered during training, resulting in a sorted list of $\Delta Q(o_i)$, which is denoted as $L_{\Delta Q(o_i)}$. Then, we set $T$ by splitting $L_{\Delta Q(o_i)}$ in terms of the index. For example, if we want to prune $T_m\%$ messages, we set $T=L_{\Delta Q(o_i)}[K \times T_m\%]$. We do not split $L_{\Delta Q(o_i)}$ in terms of the value because $\Delta Q(o_i)$ usually has a non-uniform distribution. The advantage of a fixed $T$ is that the actual number of the pruned messages is ensured to be close to the desired $T_m\%$. Besides, this method is friendly to a large $K$.

To calculate the dynamic $T$, we adopt the exponential moving average technique to update $T$ as follows:
\begin{eqnarray}
T_{t} = (1-\beta) T_{t-1} + \beta ( \hspace{0.3em} Q_t(\langle o_{i}, a_{i}^{C} \rangle, \langle \vec{o}_{-i}, \vec{a}_{-i}^{C} \rangle) \nonumber \\
- Q_t(\langle o_{i}, a_{i}^{I} \rangle, \langle \vec{o}_{-i}, \vec{a}_{-i}^{C} \rangle) \hspace{0.3em} ) \label{equ:T}
\end{eqnarray}
where $\beta$ is a coefficient for discounting older $T$, and the subscript $t$ represents the training timestep. We test some $\beta$ in $[0.6,0.9]$, and they all work well. The advantage of a dynamic $T$ is that $Y(o_i)$ becomes an adaptive training label even for the same observation $o_i$. This is very important for the dynamically changing environments because $T$ and $Y(o_i)$ can quickly adapt to these environments.

\subsection{Apply Gating to Previous Methods}
Although we introduce the proposed gating mechanism (i.e., the message pruning method) based on the basic ACML model, it is not specifically tailored to any specific DRL architecture. There are two directions to extend our method. On the one hand, we can apply the gating mechanism to previous methods such as CommNet and AMP, so that the resulting Gated-CommNet and Gated-AMP can also be applied in limited bandwidth setting. On the other hand, the basic Gated-ACML adopts a fully-connected MessageCoordinatorNet to process the received messages, so we can improve the fully-connected network with more advanced networks, such as the mean network in CommNet, the attention network in AMP and the BiRNN network in BiCNet \cite{peng2017multiagent}. We mainly explore the former extension in the experiments because we focus on message pruning before sending the message rather than message processing after receiving the message.

\section{Experimental Evaluations}
\subsection{Environments}
\begin{figure}[!htb]
	\centering
	\begin{subfigure}{.33\columnwidth}
		\centering
		\includegraphics[width=.95\textwidth]{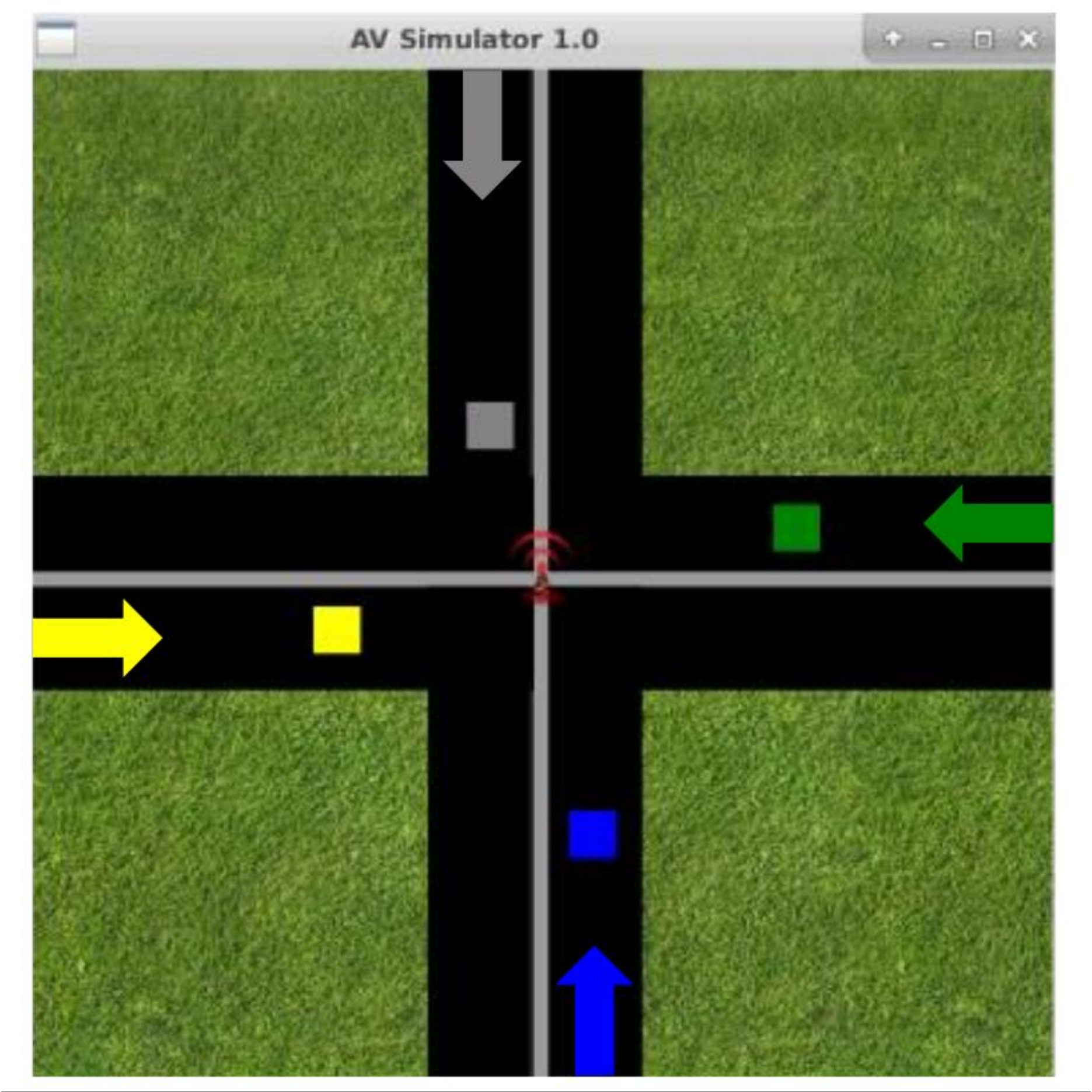}
		\caption{4 cars.}
		\label{fig:Traffic4}
	\end{subfigure}
	\begin{subfigure}{.33\columnwidth}
		\centering
		\includegraphics[width=.95\textwidth]{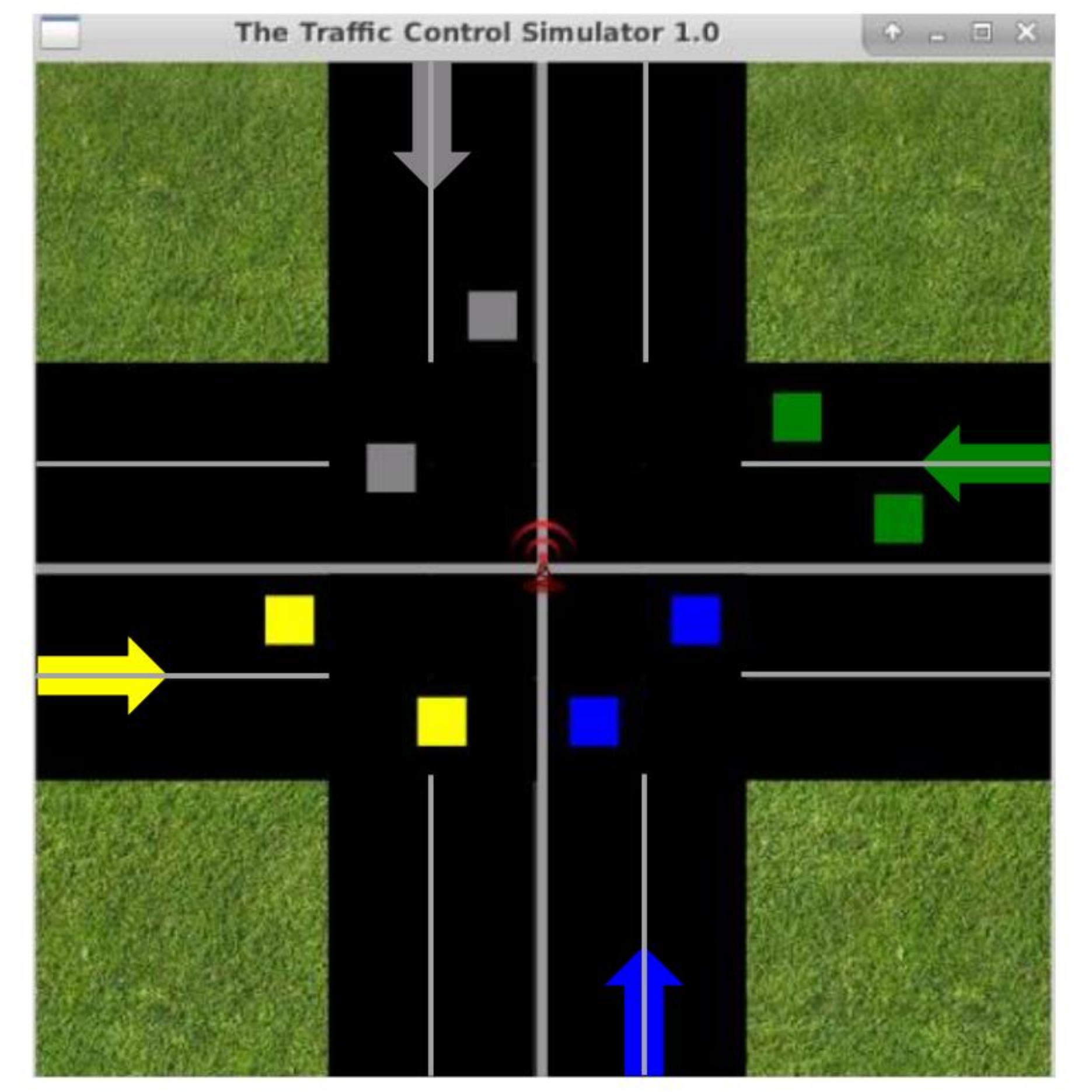}
		\caption{8 cars.}
		\label{fig:Traffic8}
	\end{subfigure}
	\begin{subfigure}{.32\columnwidth}
		\centering
		\includegraphics[width=.95\textwidth]{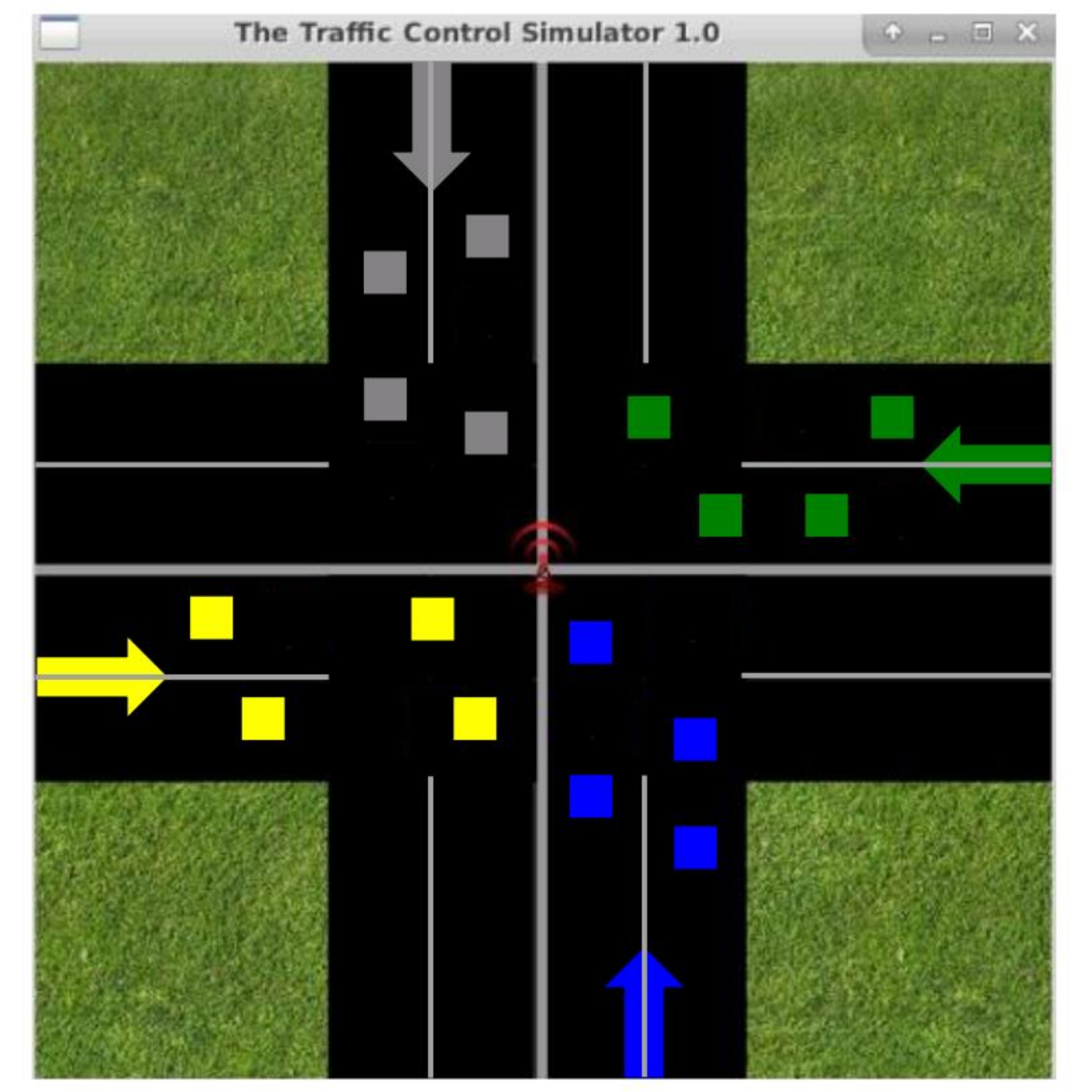}
		\caption{16 cars.}
		\label{fig:Traffic16}
	\end{subfigure}
	\begin{subfigure}{.33\columnwidth}
		\centering
		\includegraphics[width=.95\textwidth]{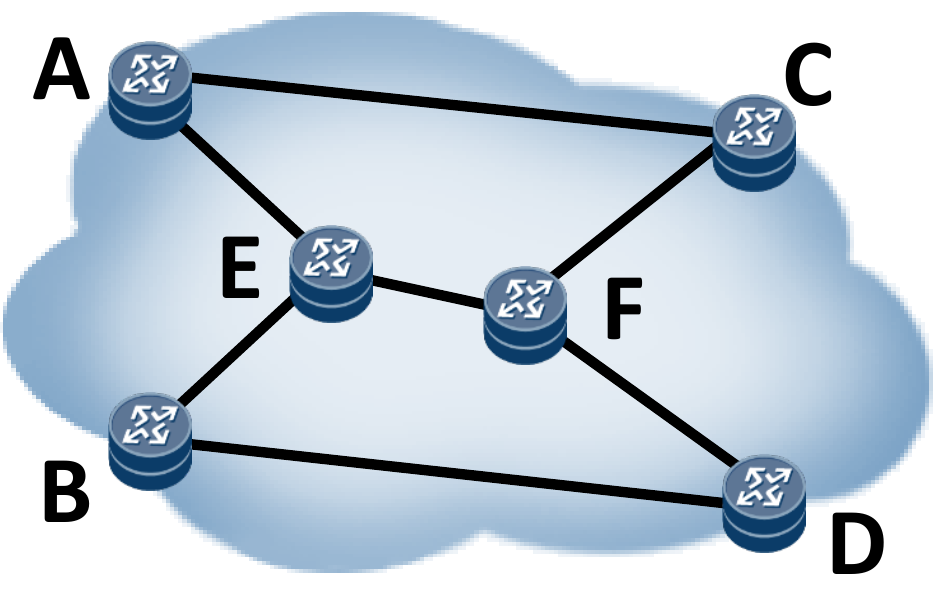}
		\caption{6 routers along with 4 paths.}
		\label{fig:RoutingSmall}
	\end{subfigure}
	\begin{subfigure}{.33\columnwidth}
		\centering
		\includegraphics[width=.95\textwidth]{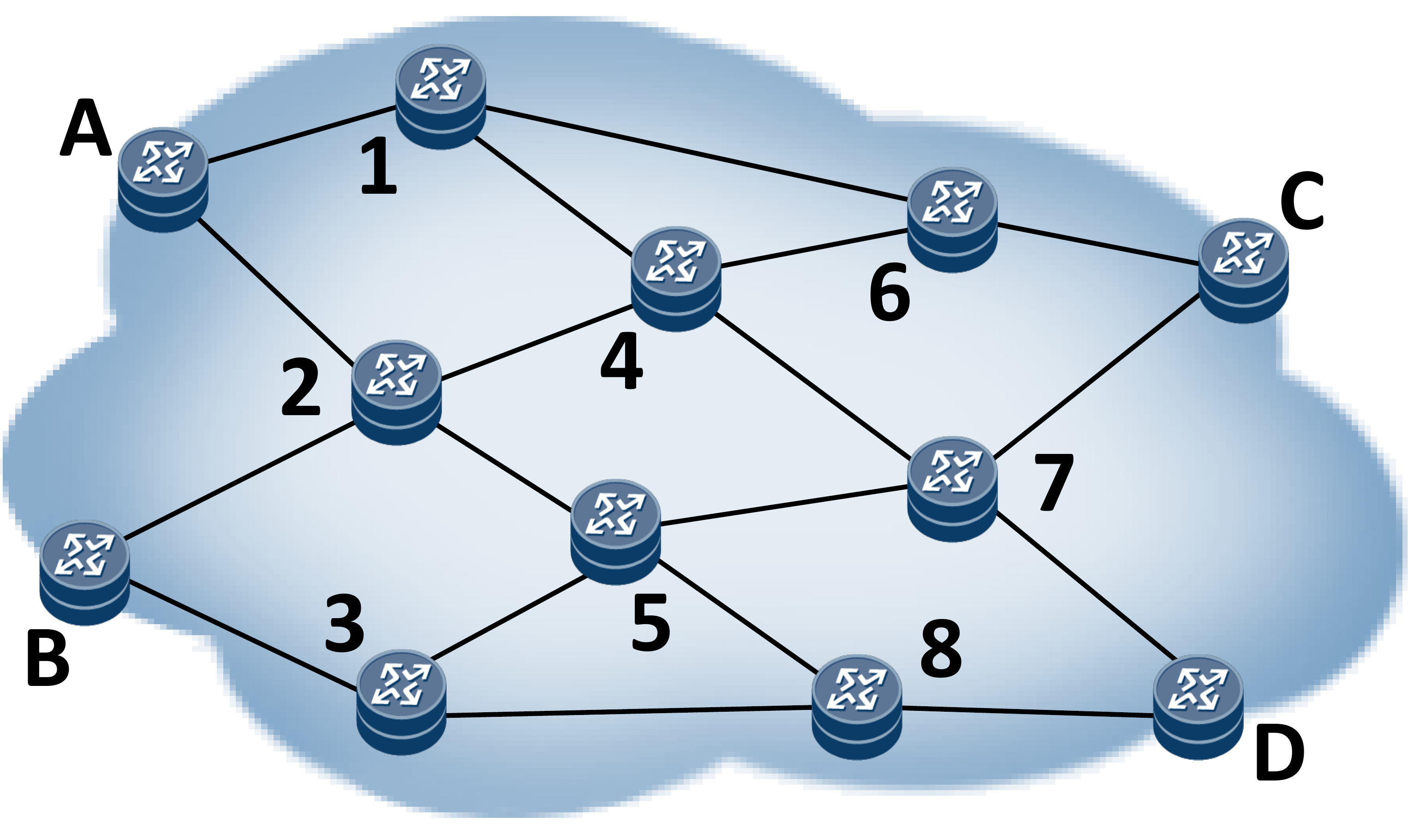}
		\caption{12 routers along with 20 paths.}
		\label{fig:RoutingLarge}
	\end{subfigure}
	\begin{subfigure}{.32\columnwidth}
		\centering
		\includegraphics[width=.95\textwidth]{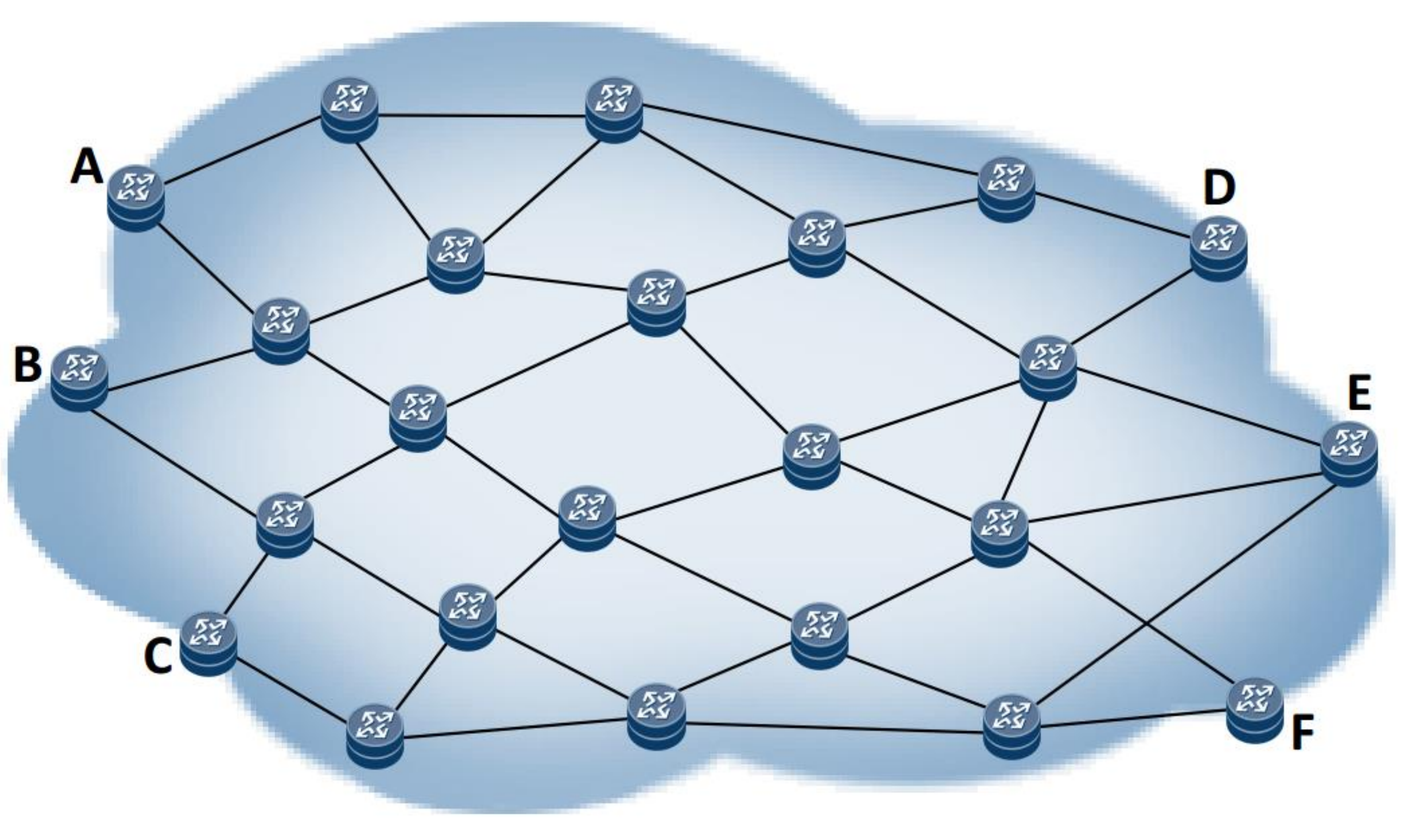}
		\caption{24 routers along with 128 paths!}
		\label{fig:RoutingHuge}
	\end{subfigure}
	\begin{subfigure}{.35\columnwidth}
		\centering
		\includegraphics[width=.95\textwidth]{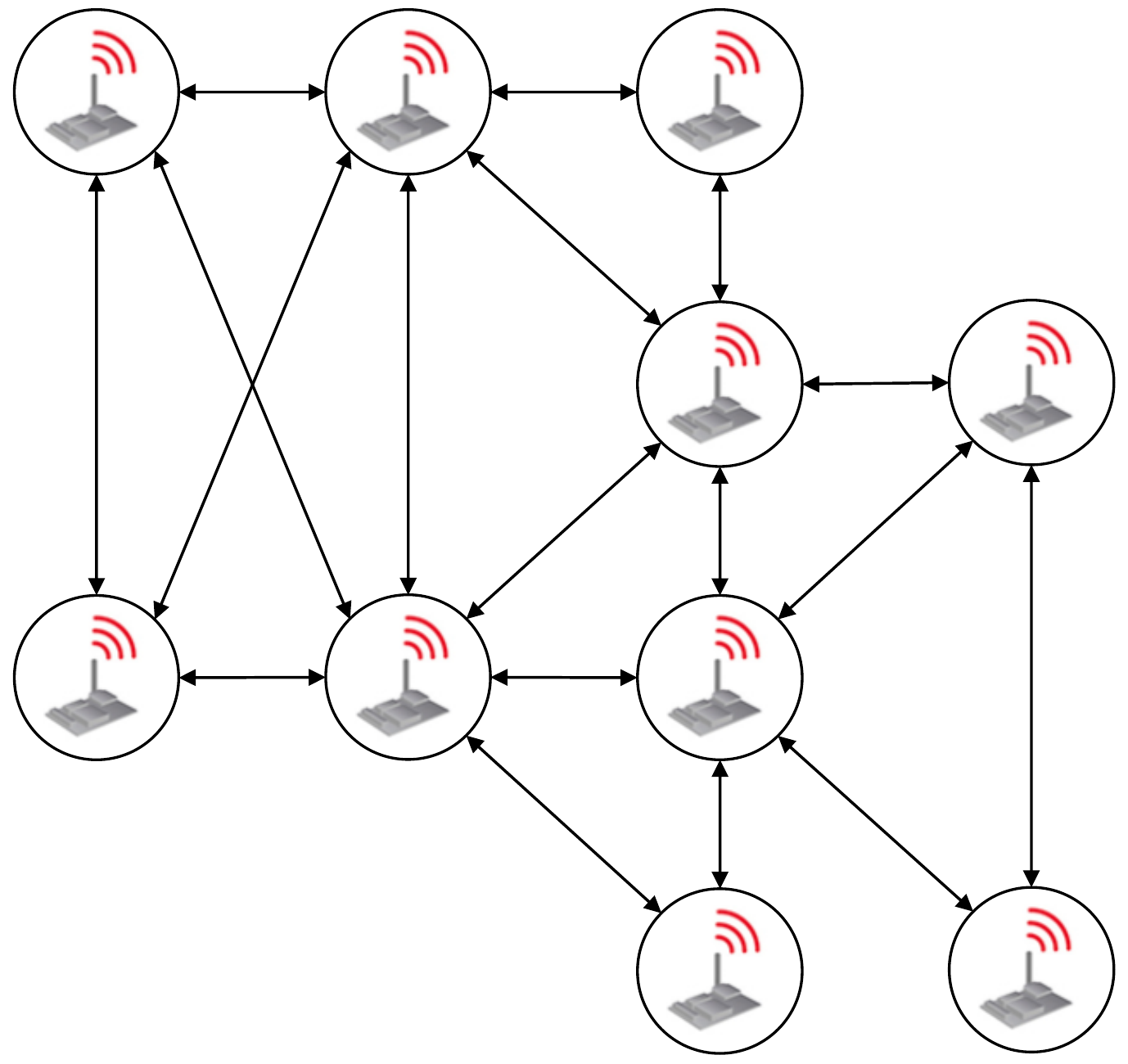}
		\caption{10 access points with 35 wireless links.}
		\label{fig:IoT10}
	\end{subfigure}
	\begin{subfigure}{.64\columnwidth}
		\centering
		\includegraphics[width=.95\textwidth]{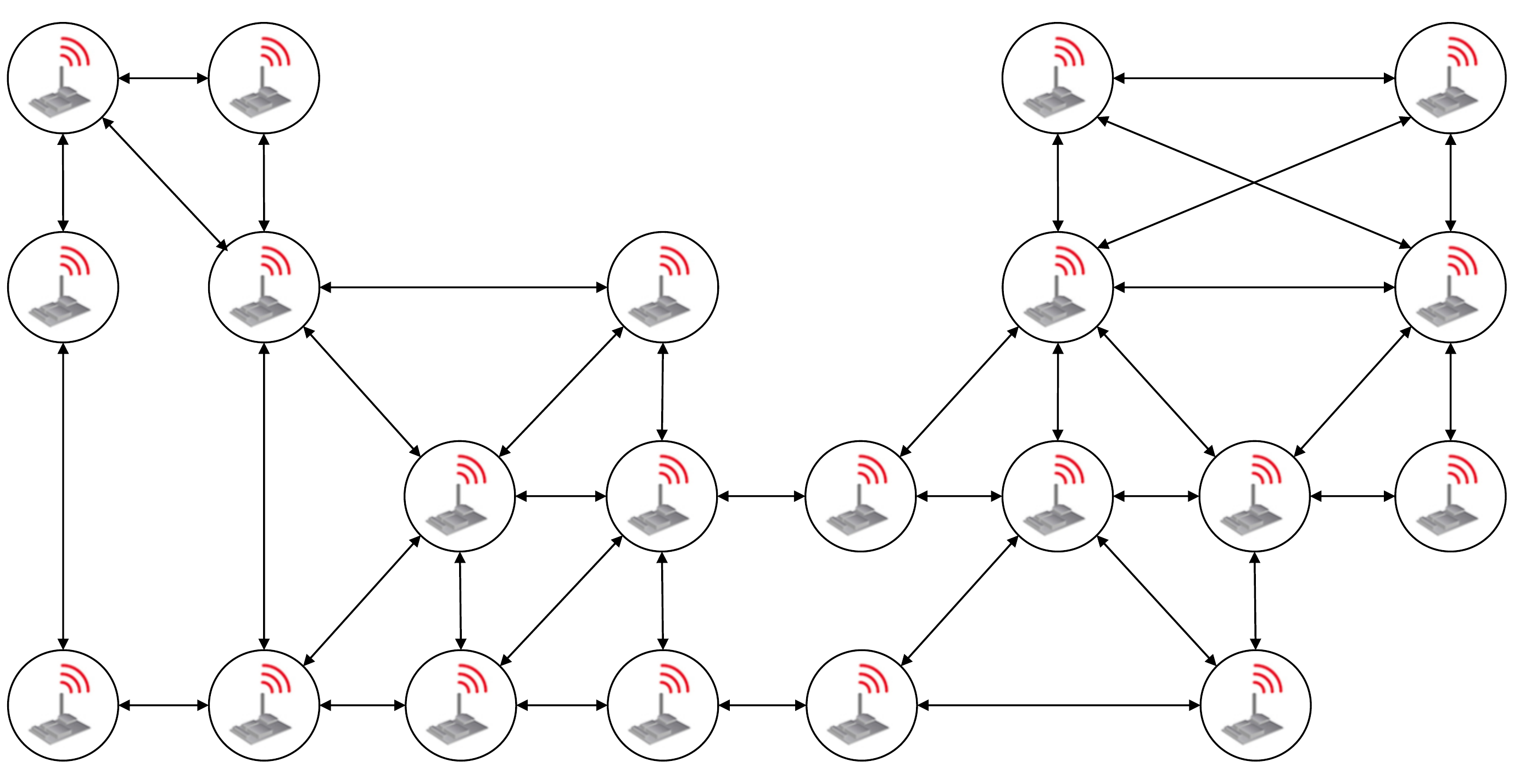}
		\caption{21 access points along with 70 wireless links!}
		\label{fig:IoT21}
	\end{subfigure}
	\caption{The evaluation tasks. (a-c) The simple, moderate and complex traffic control scenarios. (d-f) The simple, moderate and complex packet routing scenarios. (g-h) The simple and complex access point configuration scenarios.}
	\label{fig:EvaluationEnvironments}
\end{figure}
\textbf{Traffic Control.} As shown in Figure \ref{fig:Traffic4}, the cars are driving on the road. The car collision occurs when the locations of two cars are overlapped, but it does not affect the simulation except for the reward these cars receiving. The cars are controlled by our method, and they try to learn a good driving policy to cooperatively drive through the junction. The simulation is terminated after 100 steps or when all cars successfully exit the junction. For each car, the \emph{\textbf{observation}} encodes its current location and assigned route number. The \emph{\textbf{action}} is a real number $a \in (0, 1)$, which indicates how far to move ahead the car on its route. For the reward, each car gets a reward $r^{\tau}_{time}=-0.1\tau$ at each timestep to discourage a traffic jam, where $\tau$ is the total timesteps since the car appeared in the simulator; in addition, a car collision incurs a penalty $r_{coll}=-10.0$ on the received reward, while an additional reward $r_{exit}=30.0$ will be given if the car successfully exits the junction; thus, the \emph{\textbf{total reward}} at time $t$ is: $r(t)=\sum_{i=1}^{N^{t}} r^{\tau_{i}}_{time} + C^{t}r_{coll} + E^{t}r_{exit}$, where $N^{t}$, $C^{t}$ and $E^{t}$ are the numbers of cars present, car collisions and cars exiting at timestep $t$, respectively.

\textbf{Packet Routing.} As shown in Figure \ref{fig:RoutingSmall}, the edge router has an aggregated flow that should be transmitted to other edge routers through available paths (e.g., $B$ is set to transmit flow to $D$, and the available paths are $BEFD$ and $BD$). Each path is made up of several links, and each link has a \emph{link utilization}, which equals to the ratio of the current flow on this link to the maximum capacity of this link. The routers are controlled by our algorithms, and they try to learn a good flow transmission policy to minimize the \emph{Maximum Link Utilization in the whole network (MLU)}. The intuition behind this objective is that high link utilization is undesirable for dealing with bursty traffic. For each router, the \textbf{\emph{observation}} includes the flow demands in its buffers, the estimated link utilization of its direct links during the last ten steps, the average link utilization of its direct links during the last control cycle, and the latest action taken by the router. The \textbf{\emph{action}} is the flow rate assigned to each available path. The \textbf{\emph{reward}} is $1-MLU$ because we want to minimize \emph{MLU}

\begin{table*}[!htb]
	\centering
	\caption{The average results of 10 experiments on traffic control tasks. For models named as Gated-*, dynamic thresholds with $\beta = 0.8$ are used. The ``delay'' indicates the timesteps to complete the simulation. The ``\# C'' indicates the number of collisions.}
	\label{tab:RESULT_TrafficControl_DynamicT}
	\resizebox{0.98\textwidth}{!}{%
		\begin{tabular}{lrrrrrrrrrrrrrr}
			\hline\noalign{\smallskip}
			& \multicolumn{4}{c}{\textbf{Simple Traffic Control (4 cars)}} & \multicolumn{4}{c}{\textbf{Moderate Traffic Control (8 cars)}} & \multicolumn{4}{c}{\textbf{Complex Traffic Control (16 cars)}} \\
			& reward & delay & \# C & message & reward & delay & \# C & message & reward & delay & \# C & message \\
			\noalign{\smallskip}\hline\noalign{\smallskip}\noalign{\smallskip}
			CommNet & -129.2 & 45.6 & 2.3 & 100.0\% & -573.4 & 61.6 & 6.8 & 100.0\% & -4278.8 & 82.1 & 16.5 & 100.0\% \\
			AMP & -97.1 & 41.8 & \bf 1.2 & 100.0\% & -1950.6 & 100.0 & \bf 0.9 & 100.0\% & -6391.5 & 100.0 & \bf 2.1 & 100.0\% \\
			ACML & \bf -37.6 & \bf 31.2 & 1.9 & 100.0\% & \bf -103.5 & \bf 43.1 & 1.7 & 100.0\% & \bf -2824.9 & \bf 66.4 & 7.2 & 100.0\% \\
			\hline
			ACML-mean & -32.9 & 29.3 & 2.0 & 100.0\% & -96.1 & 40.2 & 3.9 & 100.0\% & -2661.5 & 61.4 & 9.4 & 100.0\% \\
			ACML-attention & -24.5 & 28.8 & 1.6 & 100.0\% & -91.8 & 39.6 & 1.3 & 100.0\% & -2359.7 & 52.9 & 6.8 & 100.0\% \\
			\noalign{\smallskip}\hline\noalign{\smallskip}\noalign{\smallskip}
			Gated-CommNet & -88.4 & 41.1 & 1.7 & 22.8\% & -476.7 & 47.2 & 2.5 & 34.3\% & -3529.4 & 73.0 & 15.7 & 29.3\% \\
			Gated-AMP & -59.9 & 37.1 & \bf 1.3 & \bf 18.6\% & -988.6 & 65.3 & \bf 1.6 & \bf 23.7\% & -2870.5 & 65.5 & \bf 7.9 & \bf 19.1\% \\
			Gated-ACML & \bf -14.6 & \bf 21.0 & 2.4 & 23.9\% & \bf -69.4 & \bf 32.3 & 2.1 & 29.8\% & \bf -2101.1 & \bf 48.6 & 11.3 & 25.8\% \\
			\hline
			ATOC & -19.7 & 25.9 & 1.9 & 37.3\% & -77.5 & 35.6 & 2.4 & 63.7\% & -2481.2 & 54.8 & 14.9 & \emph{112.5\%} \\
			\noalign{\smallskip}\hline
		\end{tabular}
	}
\end{table*}

\begin{table*}[!thb]
	\centering
	\caption{The average results of 10 experiments on packet routing and wifi access point configuration tasks. For models named as Gated-*, we adopt dynamic thresholds with $\beta = 0.8$. The ``WAPC.'' is the abbreviation of Wifi Access Point Configuration.}
	\label{tab:RESULT_PacketRouting_DynamicT}
	\resizebox{0.98\textwidth}{!}{%
		\begin{tabular}{lrrrrrrr|rrrrr}
			\hline\noalign{\smallskip}
			& \multicolumn{2}{c}{\textbf{Simple Routing}} & \multicolumn{2}{c}{\textbf{Moderate Routing}} & \multicolumn{2}{c}{\textbf{Complex Routing}} &&& \multicolumn{2}{c}{\textbf{Simple WAPC.}} & \multicolumn{2}{c}{\textbf{Complex WAPC.}} \\
			& reward & message & reward & message & reward & message &&& reward & message & reward & message \\
			\noalign{\smallskip}\hline\noalign{\smallskip}\noalign{\smallskip}
			CommNet & 0.264 & 100.0\% & 0.164 & 100.0\% & - & 100.0\% &&& 0.652 & 100.0\% & 0.441 & 100.0\% \\
			AMP & 0.266 & 100.0\% & 0.185 & 100.0\% & - & 100.0\% &&& 0.627 & 100.0\% & 0.418 & 100.0\% \\
			ACML & \bf 0.317 & 100.0\% & \bf 0.263 & 100.0\% & - & 100.0\% &&& \bf 0.665 & 100.0\% & \bf 0.480 & 100.0\% \\
			\hline
			ACML-mean & 0.321 & 100.0\% & 0.267 & 100.0\% & - & 100.0\% &&& 0.673 & 100.0\% & 0.493 & 100.0\% \\
			ACML-attention & 0.329 & 100.0\% & 0.271 & 100.0\% & - & 100.0\% &&& 0.689 & 100.0\% & 0.506 & 100.0\% \\
			\noalign{\smallskip}\hline\noalign{\smallskip}\noalign{\smallskip}
			Gated-CommNet & 0.232 & 35.2\% & 0.144 & \bf 21.7\% & - & 19.8\% &&& 0.595 & 53.1\% & 0.386 & 41.8\% \\
			Gated-AMP & 0.241 & 46.7\% & 0.170 & 35.0\% & - & 81.7\% &&& 0.539 & 57.2\% & 0.350 & \bf 32.3\% \\
			Gated-ACML & 0.288 & \bf 33.6\% & \bf 0.239 & 27.9\% & - & 22.6\% &&& \bf 0.610 & \bf 41.9\% & \bf 0.411 & 37.7\% \\
			\hline
			ATOC & \bf 0.297 & 73.7\% & 0.102 & \emph{104.6\%} & - & \emph{326.1\%} &&& 0.418 & \emph{136.5\%} & 0.231 & \emph{393.4\%} \\
			\noalign{\smallskip}\hline
		\end{tabular}
	}
\end{table*}

\textbf{Wifi Access Point Configuration.} The cornerstone of any wireless network is the sensor and access point (AP). The primary job of an AP is to broadcast a wireless signal that sensors can detect and tune into. It is tedious to configure the power of APs, and the AP behaviors differ greatly with various scenarios. The current optimization is highly depending on human expertise, which fails to handle dynamic environments. In the tasks shown in Figure \ref{fig:IoT10}, the APs are controlled by our method, and they try to learn a good power configuration policy to maximize the Received Signal Strength Indicator(RSSI). In general, larger RSSI indicates better signal quality. For each AP, the \textbf{\emph{observation}} includes radio frequency, bandwidth, the rate of package loss, the number of band, the current number of users in one specific band, the number of download bytes in ten seconds, the upload coordinate speed (Mbps), the download coordinate speed and the latency. The \textbf{\emph{action}} is the power value ranging from 10.0 to 30.0. The \textbf{\emph{reward}} is RSSI and the goal is to maximize accumulated RSSI.

\subsection{Results}
\textbf{Results without Message Pruning.} We firstly evaluate the ACML model. For traffic control, ACML outperforms CommNet and AMP as shown by the upper part of Table \ref{tab:RESULT_TrafficControl_DynamicT}. The reason is that ACML has found better tradeoffs between small delay and small collision. Please recall the settings: the reward penalty of a great delay is much larger than that of a collision. We notice that ACML drives the cars with a relatively larger speed to complete the simulation with the smallest delay (i.e., 31.2, 43.1 and 66.4 for three scenarios, respectively). Besides, although its collision is not the smallest, ACML manages to achieve relatively smaller collisions (i.e., 1.9, 1.7 and 7.2 for three scenarios), since avoiding collision is also in favor of the total reward. In contrast, AMP has a great delay and CommNet has a great collision, which makes their total rewards unsatisfactory.

We further extend ACML with the average message operation in CommNet and the attention message operation in AMP. The resulting models are named as ACML-mean and ACML-attention, respectively. Both ACML-mean and ACML-attention improve ACML by a clear margin, and they perform much better than CommNet and AMP as shown in Table \ref{tab:RESULT_TrafficControl_DynamicT}. The results imply that ACML is a general model, and it can be easily extended with advanced message processing operations to achieve better results.

In addition, the results of packet routing and wifi access point configuration shown at the upper part of Table \ref{tab:RESULT_PacketRouting_DynamicT} can be analyzed similarly to demonstrate that the performance of ACML is consistent in several multi-agent scenarios.

\textbf{Results of Dynamic Threshold Message Pruning.} For packet routing and access point configuration scenarios, as shown at the lower part of Table \ref{tab:RESULT_PacketRouting_DynamicT}, Gated-ACML can prune a lot of messages (e.g., about $70\% \approx 1 - 33.6\%$) with little damage to the reward (e.g., about $10\% \approx 1 - 0.288/0.317$). It implies that the pruned messages are usually not beneficial enough to performance, which approves the effectiveness of our design. However, none of the tested methods can handle the complex routing task! In this case, they cannot provide correct label for the auxiliary loss function in Equation \ref{equ:LossFunctionGating}, so the message quantity is almost random. We leave the results here to show the limitation of our method and others.

\begin{figure}[!thb]
	\centering
	\begin{subfigure}{.46\columnwidth}
		\centering
		\includegraphics[width=.98\textwidth]{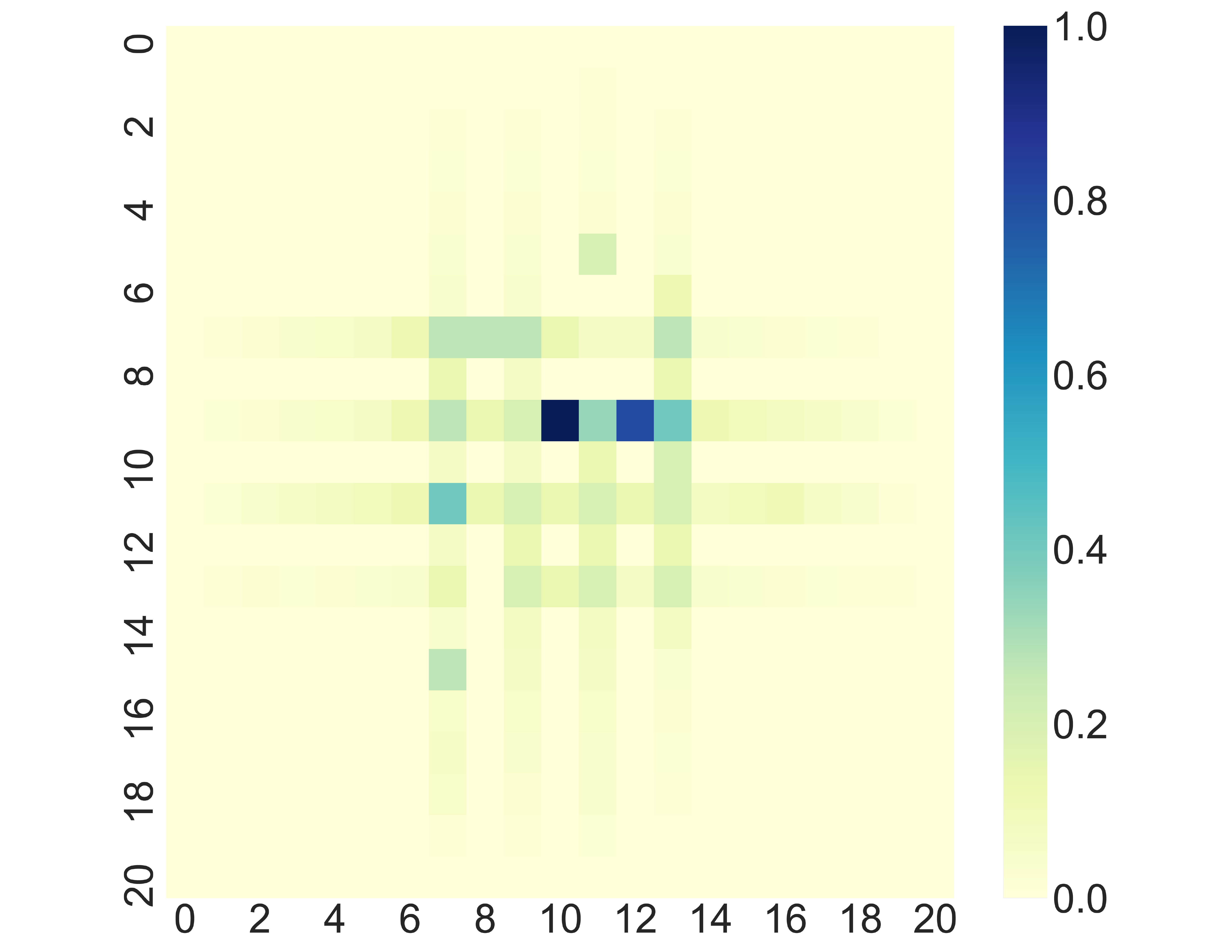}
		\caption{For moderate traffic.}
		\label{fig:RESULT_Traffic8_MessageHeatmap}
	\end{subfigure}
	\begin{subfigure}{.46\columnwidth}
		\centering
		\includegraphics[width=.98\textwidth]{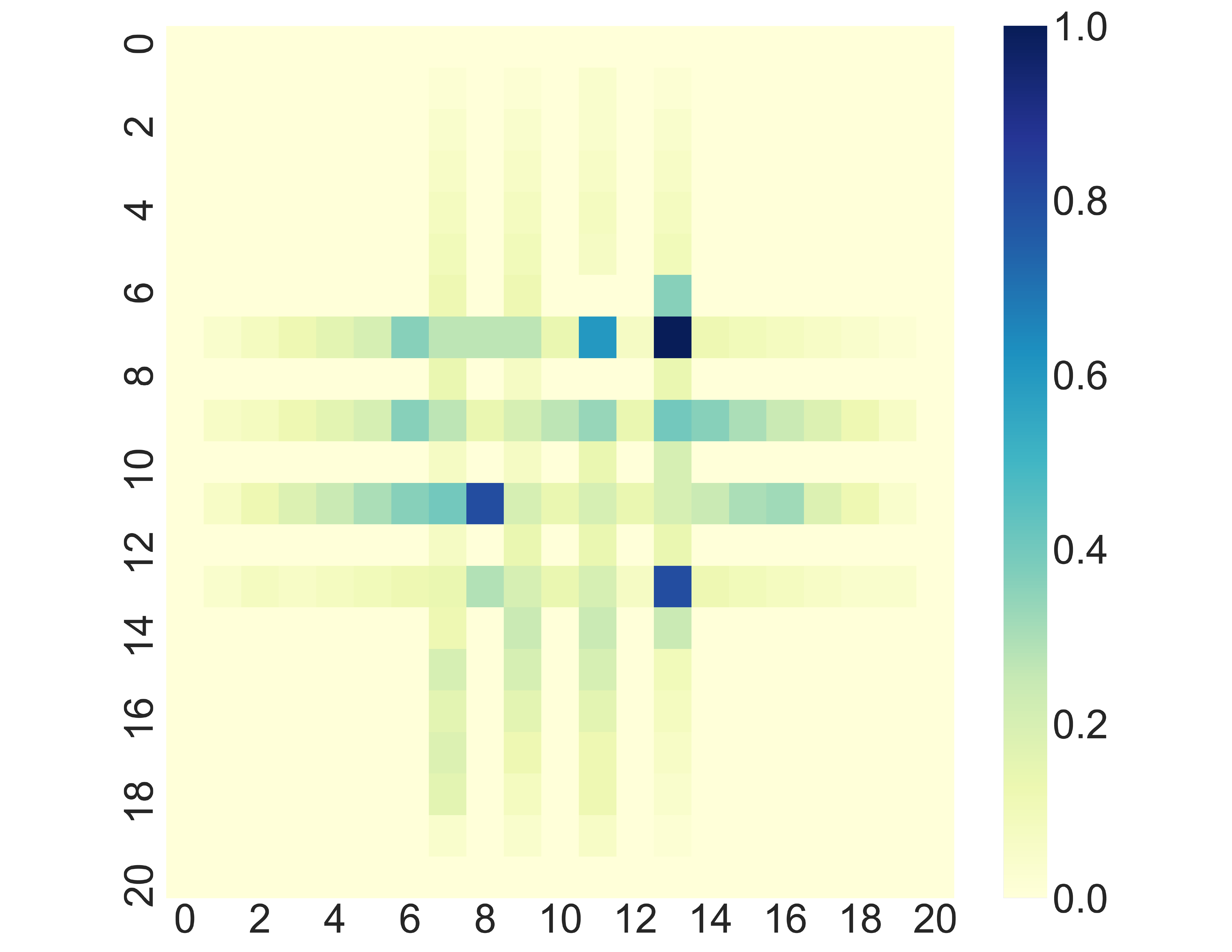}
		\caption{For complex traffic.}
		\label{fig:RESULT_Traffic16_MessageHeatmap}
	\end{subfigure}
	\caption{The message distribution of traffic control tasks.}
	\label{fig:RESULT_Traffic_MessageHeatmap}
\end{figure}

For traffic control scenarios, as shown at the lower part of Table \ref{tab:RESULT_TrafficControl_DynamicT}, Gated-ACML emits much fewer messages than ACML, which is expected. What surprises us is that Gated-ACML even obtains more rewards than ACML. This is very appealing in the real-world applications, since we can get more rewards with less message exchange and less resource consumption. We find that this phenomenon is closely related to message redundance. Figure \ref{fig:RESULT_Traffic_MessageHeatmap} shows the message distribution \footnote{We firstly discretize the 2D continuous plane into a 21-by-21 discrete plane. Then, we count the number of messages in each discrete cell. Finally, we normalize these numbers.} generated by Gated-ACML. As can be observed, the messages are mostly distributed near the junction. We hypothesize that the messages near the junction are critical, while the messages far away from the junction are redundant. Because Gated-ACML has pruned the redundant messages, the agents are rarely disturbed, so they are more adaptive: their speed gradually decreases as they are approaching the junction, while their speed gradually increases as they are moving away from the junction. In contrast, ACML does not have the ability to prune the redundant messages, so the agents are badly disturbed, and they are very conservative: the cars are too cautious to drive with a large speed even when they are far away from the junction. Consequently, the delay of ACML is larger than that of Gated-ACML as shown in Table \ref{tab:RESULT_TrafficControl_DynamicT}. Moreover, the message distribution is consistent with human experience. Because car collision easily occurs at the junction, keeping these messages is very important for avoiding car collision.

For all evaluation scenarios, Gated-CommNet and Gated-AMP reveal the same trend as Gated-ACML. It means that the gating mechanism with dynamic threshold is generally applicable to several DRL methods and multi-agent tasks.

\begin{table}[!thb]
	\centering
	\newcommand{\tabincell}[2]{\begin{tabular}{@{}#1@{}}#2\end{tabular}}
	\caption{The results of Gated-ACML in packet routing scenarios. We adopt a fixed threshold $T=L_{\Delta Q_{(o_i)}}[K \times T_m\%]$.}
	\label{tab:RESULT_PacketRouting_FixedT}
	\resizebox{0.99\columnwidth}{!}{%
		\begin{tabular}{lrrrr}
			\hline\noalign{\smallskip}
			& \multicolumn{2}{c}{Simple Routing} & \multicolumn{2}{c}{Moderate Routing} \\
			$T_m$\% & \tabincell{c}{\emph{pruned} \\ message} & \tabincell{c}{reward \\ \emph{decrease}} & \tabincell{c}{\emph{pruned} \\ message} & \tabincell{c}{reward \\ \emph{decrease}} \\
			\noalign{\smallskip}\hline\noalign{\smallskip}
			10.0\% & 12.19\% & \bf -8.46\% & 11.60\% & \bf -7.03\% \\
			20.0\% & 24.07\% & \bf -13.59\% & 22.77\% & \bf -12.14\% \\
			30.0\% & 27.65\% & \bf -4.88\% & 29.98\% & \bf -3.25\% \\
			\hline
			70.0\% & 66.73\% & 9.27\% & 68.54\% & 10.06\% \\
			80.0\% & \bf 79.14\% & \bf 14.01\% & 76.81\% & 13.25\% \\
			90.0\% & 87.22\% & 18.60\% & 85.11\% & 19.50\% \\
			100.0\% & 100.00\% & 59.35\% & 100.00\% & 65.42\% \\
			\hline\noalign{\smallskip}
		\end{tabular}
	}
\end{table}

We also modify ATOC to make it suitable for continuous action and heterogeneous agents. The results are shown at the last row of Table \ref{tab:RESULT_TrafficControl_DynamicT} and \ref{tab:RESULT_PacketRouting_DynamicT}. ATOC works well in traffic control tasks, but performs badly in routing and wifi tasks. The main reason is that ATOC decomposes all agents into small agent groups by distance between agents. It is in favor of tasks defined on 2D plane (e.g., traffic control). However, the routers or APs are entangled with each other by other factors but not distance between them. Critically, the results show that ATOC could hardly be used in limited bandwidth setting: it emits much more messages than our methods (e.g., about tenfold messages of ours in complex wifi scenario). This is because agents in ATOC can belong to many agent groups, and thus generate many messages. The results show the urgent demand of principled message pruning methods.

\textbf{Results of Fixed Threshold Message Pruning.} The results are shown in Table \ref{tab:RESULT_PacketRouting_FixedT}. It can be noticed that the number of pruned messages is close to the desired $T_m\%$ for both routing scenarios. It is the advantage of fixed threshold as mentioned before, but it has not been implemented by any previous methods as far as we know. In addition, as shown at the upper part of the table, when the quantity of \emph{pruned} messages is smaller than 30\%, the reward \emph{decrease} is a negative value, which means that the reward is increased actually. This is because some messages are extremely noisy in the routing system, and pruning these messages will be beneficial for the performance. Please note that similar phenomena have been observed in traffic control scenarios. However, as we are pruning more messages, the reward decrease is becoming larger and larger as shown at the lower part of the table, especially when all messages are pruned. It indicates that the remaining messages are very critical for maintaining the performance, and Gated-ACML has learnt to share these important messages with other agents. In contrast, even if a large number of messages (e.g., $79.14\%$) have been pruned, the reward decrease is not so great (e.g., $14.01\%$). Again, it implies that the pruned messages are less beneficial to the performance, and Gated-ACML with fixed threshold has mastered an effective strategy to prune these messages.


\section{Conclusions}
We have proposed a gating mechanism, which consists of several key designs like auxiliary task with appropriate training signal, dynamic and fixed thresholds, to address the limited bandwidth that has been largely ignored by previous DRL methods. The gating mechanism prunes less beneficial messages in an adaptive manner, so that the performance can be maintained or even improved with much fewer messages. Furthermore, it is applicable to several previous methods and multi-agent scenarios with good performance. To the best of our knowledge, it is the first method to achieve these in the multi-agent reinforcement learning community. 

\section{Acknowledgments}
The authors would like to thank the anonymous reviewers for their comments. This work was supported by the National Natural Science Foundation of China under Grant No.61872397. The contact author is Zhen Xiao.

\bibliography{2312.aaai20} 

\begin{thebibliography}{}

\bibitem[\protect\citeauthoryear{Becker \bgroup et al\mbox.\egroup
  }{2009}]{becker2009analyzing}
Becker, R.; Carlin, A.; Lesser, V.; and Zilberstein, S.
\newblock 2009.
\newblock Analyzing myopic approaches for multi-agent communication.
\newblock {\em Computational Intelligence} 25(1):31--50.

\bibitem[\protect\citeauthoryear{Bernstein \bgroup et al\mbox.\egroup
  }{2002}]{bernstein2002complexity}
Bernstein, D.~S.; Givan, R.; Immerman, N.; and Zilberstein, S.
\newblock 2002.
\newblock The complexity of decentralized control of markov decision processes.
\newblock {\em Mathematics of operations research} 27(4):819--840.

\bibitem[\protect\citeauthoryear{Chu and Ye}{2017}]{Chu2017Parameter}
Chu, X., and Ye, H.
\newblock 2017.
\newblock Parameter sharing deep deterministic policy gradient for cooperative
  multi-agent reinforcement learning.
\newblock {\em arXiv preprint arXiv:1710.00336}.

\bibitem[\protect\citeauthoryear{Foerster \bgroup et al\mbox.\egroup
  }{2016}]{foerster2016learning}
Foerster, J.; Assael, I.~A.; de~Freitas, N.; and Whiteson, S.
\newblock 2016.
\newblock Learning to communicate with deep multi-agent reinforcement learning.
\newblock In {\em Advances in Neural Information Processing Systems},
  2137--2145.

\bibitem[\protect\citeauthoryear{Hernandez-Leal \bgroup et al\mbox.\egroup
  }{2017}]{hernandez2017survey}
Hernandez-Leal, P.; Kaisers, M.; Baarslag, T.; and de~Cote, E.~M.
\newblock 2017.
\newblock A survey of learning in multiagent environments: Dealing with
  non-stationarity.
\newblock {\em arXiv preprint arXiv:1707.09183}.

\bibitem[\protect\citeauthoryear{Hubara \bgroup et al\mbox.\egroup
  }{2016}]{Hubara2016Binarized}
Hubara, I.; Courbariaux, M.; Soudry, D.; El-Yaniv, R.; and Bengio, Y.
\newblock 2016.
\newblock Binarized neural networks.
\newblock In {\em Advances in neural information processing systems},
  4107--4115.

\bibitem[\protect\citeauthoryear{Jaderberg \bgroup et al\mbox.\egroup
  }{2016}]{Jaderberg2016Reinforcement}
Jaderberg, M.; Mnih, V.; Czarnecki, W.~M.; Schaul, T.; Leibo, J.~Z.; Silver,
  D.; and Kavukcuoglu, K.
\newblock 2016.
\newblock Reinforcement learning with unsupervised auxiliary tasks.
\newblock {\em arXiv preprint arXiv:1611.05397}.

\bibitem[\protect\citeauthoryear{Jiang and Lu}{2018}]{jiang2018learning}
Jiang, J., and Lu, Z.
\newblock 2018.
\newblock Learning attentional communication for multi-agent cooperation.
\newblock {\em Advances in neural information processing systems}.

\bibitem[\protect\citeauthoryear{Kilinc and Montana}{2018}]{kilinc2018multi}
Kilinc, O., and Montana, G.
\newblock 2018.
\newblock Multi-agent deep reinforcement learning with extremely noisy
  observations.
\newblock {\em arXiv preprint arXiv:1812.00922}.

\bibitem[\protect\citeauthoryear{Kim \bgroup et al\mbox.\egroup
  }{2019}]{kim2019learning}
Kim, D.; Moon, S.; Hostallero, D.; Kang, W.~J.; Lee, T.; Son, K.; and Yi, Y.
\newblock 2019.
\newblock Learning to schedule communication in multi-agent reinforcement
  learning.
\newblock {\em International Conference on Learning Representations}.

\bibitem[\protect\citeauthoryear{Kim, Cho, and Sung}{2019}]{kim2019message}
Kim, W.; Cho, M.; and Sung, Y.
\newblock 2019.
\newblock Message-dropout: An efficient training method for multi-agent deep
  reinforcement learning.
\newblock {\em Thirty-Third AAAI Conference on Artificial Intelligence}.

\bibitem[\protect\citeauthoryear{Kong \bgroup et al\mbox.\egroup
  }{2017}]{kong2017revisiting}
Kong, X.; Xin, B.; Liu, F.; and Wang, Y.
\newblock 2017.
\newblock Revisiting the master-slave architecture in multi-agent deep
  reinforcement learning.
\newblock {\em arXiv preprint arXiv:1712.07305}.

\bibitem[\protect\citeauthoryear{Lillicrap \bgroup et al\mbox.\egroup
  }{2015}]{lillicrap2015continuous}
Lillicrap, T.~P.; Hunt, J.~J.; Pritzel, A.; Heess, N.; Erez, T.; Tassa, Y.;
  Silver, D.; and Wierstra, D.
\newblock 2015.
\newblock Continuous control with deep reinforcement learning.
\newblock {\em arXiv preprint arXiv:1509.02971}.

\bibitem[\protect\citeauthoryear{Lowe \bgroup et al\mbox.\egroup
  }{2017}]{Lowe2017Multi}
Lowe, R.; Wu, Y.; Tamar, A.; Harb, J.; Abbeel, O.~P.; and Mordatch, I.
\newblock 2017.
\newblock Multi-agent actor-critic for mixed cooperative-competitive
  environments.
\newblock In {\em Advances in Neural Information Processing Systems},
  6379--6390.

\bibitem[\protect\citeauthoryear{Makhzani and Frey}{2015}]{Makhzani2015Winner}
Makhzani, A., and Frey, B.~J.
\newblock 2015.
\newblock Winner-take-all autoencoders.
\newblock In {\em Advances in Neural Information Processing Systems},
  2791--2799.

\bibitem[\protect\citeauthoryear{Mao \bgroup et al\mbox.\egroup
  }{2017}]{mao2017accnet}
Mao, H.; Gong, Z.; Ni, Y.; and Xiao, Z.
\newblock 2017.
\newblock Accnet: Actor-coordinator-critic net for "learning-to-communicate"
  with deep multi-agent reinforcement learning.
\newblock {\em arXiv preprint arXiv:1706.03235}.

\bibitem[\protect\citeauthoryear{Mao \bgroup et al\mbox.\egroup
  }{2019}]{mao2019modelling}
Mao, H.; Zhang, Z.; Xiao, Z.; and Gong, Z.
\newblock 2019.
\newblock Modelling the dynamic joint policy of teammates with attention
  multi-agent ddpg.
\newblock In {\em Proceedings of the 18th International Conference on
  Autonomous Agents and MultiAgent Systems},  1108--1116.
\newblock International Foundation for Autonomous Agents and Multiagent
  Systems.

\bibitem[\protect\citeauthoryear{Nesterov}{2012}]{Nesterov2012Efficiency}
Nesterov, Y.
\newblock 2012.
\newblock Efficiency of coordinate descent methods on huge-scale optimization
  problems.
\newblock {\em SIAM Journal on Optimization} 22(2):341--362.

\bibitem[\protect\citeauthoryear{Peng \bgroup et al\mbox.\egroup
  }{2017}]{peng2017multiagent}
Peng, P.; Yuan, Q.; Wen, Y.; Yang, Y.; Tang, Z.; Long, H.; and Wang, J.
\newblock 2017.
\newblock Multiagent bidirectionally-coordinated nets for learning to play
  starcraft combat games.
\newblock {\em arXiv preprint arXiv:1703.10069}.

\bibitem[\protect\citeauthoryear{Peng, Zhang, and Luo}{2018}]{peng2018learning}
Peng, Z.; Zhang, L.; and Luo, T.
\newblock 2018.
\newblock Learning to communicate via supervised attentional message
  processing.
\newblock In {\em Proceedings of the 31st International Conference on Computer
  Animation and Social Agents},  11--16.
\newblock ACM.

\bibitem[\protect\citeauthoryear{Roth, Simmons, and
  Veloso}{2005}]{roth2005reasoning}
Roth, M.; Simmons, R.; and Veloso, M.
\newblock 2005.
\newblock Reasoning about joint beliefs for execution-time communication
  decisions.
\newblock In {\em Proceedings of the fourth international joint conference on
  Autonomous agents and multiagent systems},  786--793.
\newblock ACM.

\bibitem[\protect\citeauthoryear{Roth, Simmons, and
  Veloso}{2006}]{roth2006communicate}
Roth, M.; Simmons, R.; and Veloso, M.
\newblock 2006.
\newblock What to communicate? execution-time decision in multi-agent pomdps.
\newblock In {\em Distributed Autonomous Robotic Systems 7}. Springer.
\newblock  177--186.

\bibitem[\protect\citeauthoryear{Silver \bgroup et al\mbox.\egroup
  }{2014}]{silver2014deterministic}
Silver, D.; Lever, G.; Heess, N.; Degris, T.; Wierstra, D.; and Riedmiller, M.
\newblock 2014.
\newblock Deterministic policy gradient algorithms.
\newblock In {\em ICML}.

\bibitem[\protect\citeauthoryear{Singh, Jain, and
  Sukhbaatar}{2019}]{singh2018learning}
Singh, A.; Jain, T.; and Sukhbaatar, S.
\newblock 2019.
\newblock Learning when to communicate at scale in multiagent cooperative and
  competitive tasks.
\newblock {\em International Conference on Learning Representations}.

\bibitem[\protect\citeauthoryear{Sukhbaatar, Fergus, and
  others}{2016}]{sukhbaatar2016learning}
Sukhbaatar, S.; Fergus, R.; et~al.
\newblock 2016.
\newblock Learning multiagent communication with backpropagation.
\newblock In {\em Advances in Neural Information Processing Systems},
  2244--2252.

\bibitem[\protect\citeauthoryear{Sutton and
  Barto}{1998}]{sutton1998introduction}
Sutton, R.~S., and Barto, A.~G.
\newblock 1998.
\newblock {\em Introduction to reinforcement learning}, volume 135.
\newblock MIT press Cambridge.

\bibitem[\protect\citeauthoryear{Wu, Zilberstein, and
  Chen}{2011}]{wu2011online}
Wu, F.; Zilberstein, S.; and Chen, X.
\newblock 2011.
\newblock Online planning for multi-agent systems with bounded communication.
\newblock {\em Artificial Intelligence} 175(2):487--511.

\bibitem[\protect\citeauthoryear{Zhang and
  Lesser}{2013}]{zhang2013coordinating}
Zhang, C., and Lesser, V.
\newblock 2013.
\newblock Coordinating multi-agent reinforcement learning with limited
  communication.
\newblock In {\em Proceedings of the 2013 international conference on
  Autonomous agents and multi-agent systems},  1101--1108.
\newblock International Foundation for Autonomous Agents and Multiagent
  Systems.

\end{thebibliography}
\bibliographystyle{aaai} 

\end{document}